\documentclass{article}

\usepackage[numbers]{natbib}

\usepackage[table]{xcolor}
\usepackage{amsfonts}
\usepackage[preprint]{neurips_2025}

\usepackage[utf8]{inputenc} 
\usepackage[T1]{fontenc}    
\usepackage{url}            
\usepackage{booktabs}       
\usepackage{amsfonts}       
\usepackage{nicefrac}       
\usepackage{microtype}      
\usepackage{xcolor}         

\usepackage{graphicx}
\usepackage{amsmath}
\usepackage{amssymb}
\usepackage{mathtools}
\usepackage{amsthm}
\usepackage{xcolor}         
\definecolor{linkColor}{rgb}{0.2,0.4,0.6}

\usepackage{framed}

\usepackage{bm}

\usepackage{arydshln}
\usepackage{color}
\usepackage{pifont}
\usepackage{enumitem}

\usepackage[utf8]{inputenc}
\usepackage{tcolorbox}
\usepackage{listings}
\tcbuselibrary{listings, skins, breakable}

\usepackage{marvosym}

\newtcolorbox{SplitPromptBox}[1][]{ enhanced, colback=blue!5!white, colbacklower=blue!3!white, colframe=violet!60!black, colbacktitle=violet!15!white, coltitle=violet!90!black, fonttitle=\bfseries, rounded corners, boxrule=1pt, left=3mm, right=3mm, top=2mm, bottom=2mm, title=#1, segmentation style={dashed, gray!70}, breakable }

\newtcolorbox{SplitPromptBoxThree}[1][]{
    enhanced,
    breakable,
    colback=blue!5!white,
    colframe=violet!60!black,
    colbacktitle=violet!15!white,
    coltitle=violet!90!black,
    fonttitle=\bfseries,
    rounded corners,
    boxrule=1pt,
    left=3mm, right=3mm, top=2mm, bottom=2mm,
    title=#1,
}

\usepackage[colorlinks=true,linkcolor=linkColor,citecolor=linkColor,filecolor=linkColor,urlcolor=linkColor]{hyperref} 

\title{PretrainZero: Reinforcement Active Pretraining}

\author{
  Xingrun Xing\textsuperscript{1,2}, Zhiyuan Fan\textsuperscript{2}, Jie Lou\textsuperscript{2\Letter},  Guoqi Li\textsuperscript{1}, Jiajun Zhang\textsuperscript{1\Letter}, Debing Zhang\textsuperscript{2} \\
  \textsuperscript{1}~Institute of Automation, Chinese Academy of Sciences \\
  \textsuperscript{2}~Xiaohongshu Inc. \\
  \texttt{loujie0822@aliyun.com, jjzhang@nlpr.ia.ac.cn}
}

\begin{document}

\maketitle

\begin{abstract}
Mimicking human behavior to actively learning from general experience and achieve artificial general intelligence has always been a human dream.
Recent reinforcement learning (RL) based large-thinking models demonstrate impressive expert-level abilities, i.e., software and math, but still rely heavily on verifiable rewards in specific domains, placing a significant bottleneck to extend the performance boundary of general reasoning capabilities. 
In this work, we propose PretrainZero, a reinforcement active learning framework built on the pretraining corpus to extend RL from domain-specific post-training to general pretraining. PretrainZero features the following characteristics:
1) \textbf{Active pretraining}: 
inspired by the active learning ability of humans, PretrainZero learns a unified reasoning policy to actively identify reasonable and informative contents from pretraining corpus, and reason to predict these contents by RL.
2) \textbf{Self-supervised learning}: 
without any verifiable labels, pretrained reward models, or supervised fine-tuning, we directly pretrain reasoners from $3\sim30$B base models on the general Wikipedia corpus using RL, significantly breaking the verification data-wall for general reasoning.
3) \textbf{Verification scaling}: 
by tackling increasingly challenging masked spans, PretrainZero substantially enhances the general reasoning abilities of pretrained base models. In reinforcement pretraining, PretrainZero improves Qwen3-4B-Base for 8.43, 5.96 and 10.60 on MMLU-Pro, SuperGPQA and math average benchmarks. In post-training, the pretrained models can also serve as reasoning foundation models for downstream RLVR tasks.

\begin{figure}[h]
\begin{center}
\centerline{\includegraphics[width=1.0\textwidth]{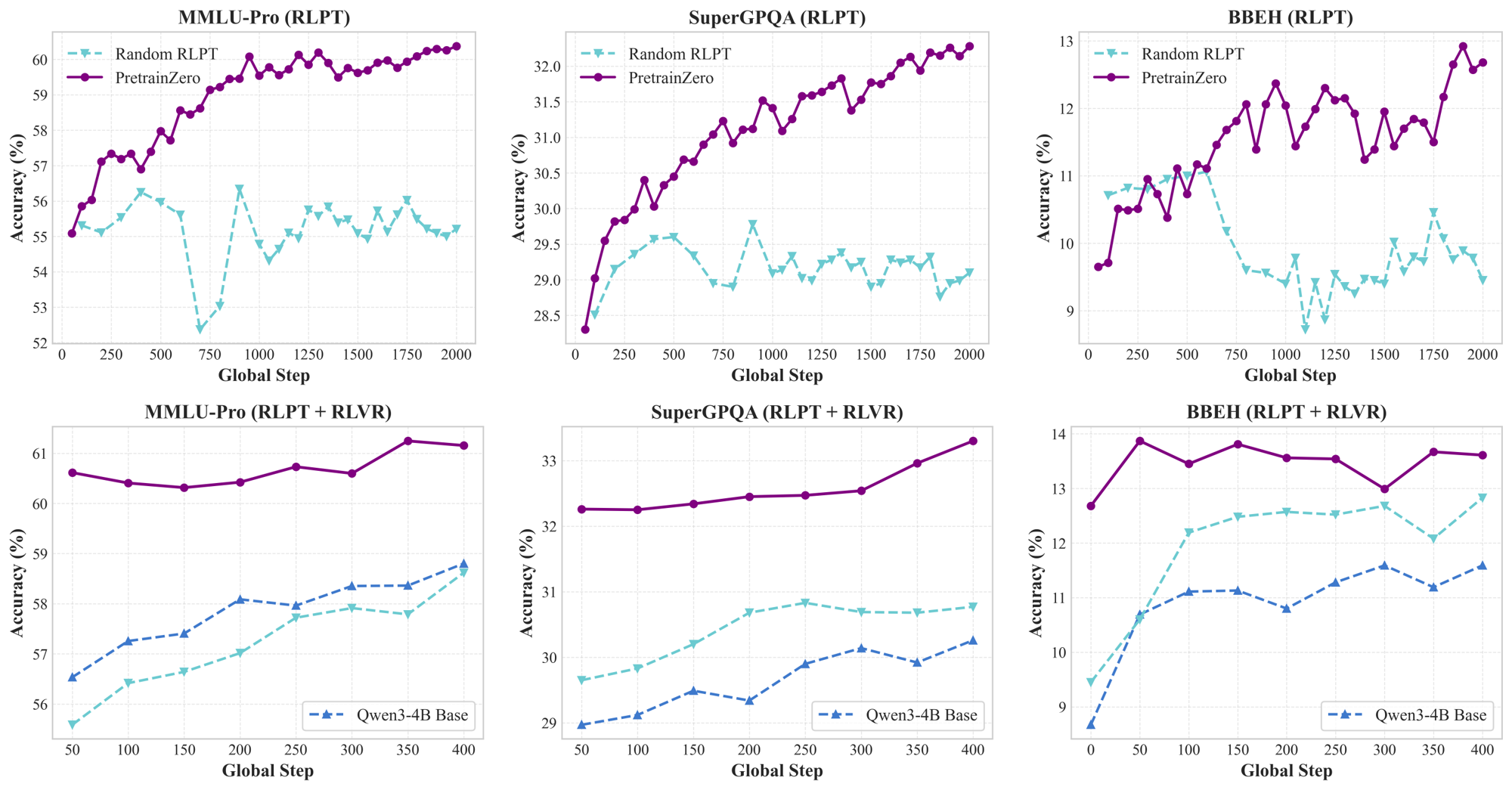}}
\vskip -0.1in 
\caption{
Reinforcement Pre-Training (RLPT) performance in pre-training and post-training stages.}
\label{f2}
\end{center}
\vskip -0.3in 
\end{figure}

\end{abstract}

\section{Introduction}

Recent large language models (LLMs) have achieved human-level expertise in specific domains, particularly through large-scale self-supervised learning in pretraining \cite{scaling:law, achiam2023gpt-4} and Reinforcement Learning (RL) \cite{deepseekr1, dapo,chu2025sft} in post-training. 
During pretraining, self-supervised learning with a fixed next-token prediction paradigm allows models to leverage large-scale, low-cost data to improve general capabilities effectively. 
In contrast, the post-training RL faces a severe data-wall: Reinforcement Learning with Verifiable Rewards (RLVR) \cite{deepseekr1,yue2504does} requires domain-specific verifiers to label training samples, and Reinforcement Learning from Human Feedback (RLHF) \cite{instructgpt, bai2022training}, relying on reward models and humans, can only train limited steps to avoid reward hacking. 
This motivates a natural direction—performing reinforcement learning \cite{dong2025reinforcement, li2025reinforcement} in a self-supervised pretraining manner \cite{gpt3}, in order to use inexpensive pretraining data to extend RLVR and overcome this data-wall.

However, formulating the self-supervised pretraining as RLVR tasks is non-trivial. Towards this goal, this work first investigates stand-alone Reinforcement Learning Pre-Training (RLPT) \cite{dong2025reinforcement} 
according to three principles: 1) Comprehensiveness: we establish both baselines including masked token prediction and next token prediction as the reasoning objective \cite{dong2025reinforcement}. 2) Full self-supervision: we exclude any additional Supervised Fine-Tuning (SFT) cold-start and reward models \cite{li2025reinforcement}. 3) Generalization: we avoid Question–Answer formats or synthetic chain-of-thought (CoT) datasets, like OmniMath \cite{omnimath}, and train directly on general-domain Wikipedia \cite{bert}. 
Experimental results demonstrate that the vanilla RLPT fails to emerge high-quality CoT: the low information density of pretraining corpus leads to inefficient learning, and the presence of noisy or incorrect tokens often causes training collapse.

This work proposes the first stand-alone RLPT method to extend RLVR on real-world pretraining corpus, termed PretrainZero. This is achieved by mimicking the human active-learning behavior \cite{yang2025active, settles2009active}: 
humans can actively learn from a broad range of experiences, selectively focusing on informative elements and unfamiliar concepts. This allows effective learning even when the underlying experiences are noisy or low in information density. In contrast, current large language models—whether through supervised or reinforcement pretraining—rely on fixed prediction patterns, such as next-token or off-policy selected masked-token prediction. These rigid learning patterns limit their efficiency \cite{berglund2309reversal} and prevent them from leveraging data as flexibly as humans do.

Inspired by this, this work proposes a reinforcement active learning framework in order to learn from real-world pretraining distributions. 
Specifically, we introduce an on-policy \textit{mask generation task} as an auxiliary active-learning objective for the \textit{masked-span prediction task}. The mask generation policy actively explores informative, verifiable, and not-yet-mastered content within the pretraining data and selects it as learning targets. Consequently, the masked-span prediction policy learns to produce CoT reasoning and recover these informative spans. 
For optimization, we formulate a min–max bilevel reinforcement learning objective, where each batch is jointly optimized using GRPO \cite{grpo} over both the mask generation and masked-span prediction tasks.
Different from unsupervised RL methods such as self-play \cite{huang2025r} and test-time scaling \cite{snell2024scaling}, PretrainZero provides a \textbf{verifiable RL scaling mechanism grounded in real data} in a self-supervising manner. This avoids the severe hallucination issues in self-play and test-time scaling, where majority voting from model-generated answers serves as supervision and ultimately leads to collapse in prolonged RL training \cite{liu2025prorl}.

As shown in Fig.~\ref{f2}, we evaluate RLPT for 2000 steps in pretraining and add general RLVR in post-training. For a Qwen3-4B-Base \cite{yang2025qwen3} model, PretrainZero consistently improves 8.43, 5.96, and 10.60 on MMLU-Pro \cite{mmlupro}, SuperGPQA \cite{supergpqa}, and math average benchmarks during reinforcement pretraining. 
After general RLVR \cite{ma2025general} in the post-training stage, these improvements remain substantial, with final improvements of 2.35, 3.04, and 2.81 on MMLU-Pro, SuperGPQA, and math average respectively.

Our contributions are summarized as follows:
\begin{itemize}[leftmargin=*]
\setlength\itemsep{0.01em}
\item We introduce PretrainZero, the first stand-alone RLPT method to operate RLVR on real-world pretraining corpus, enabling general-domain and large-scale reinforcement learning trained directly from base models using only pretraining data as grounding.
\item We propose the reinforcement active pretraining mechanism inspired by human active learning. The introduced mask-generation objective enables the model to anticipate what information should be learned actively, ensuring effective training under low–information-density pretraining corpus.
\item We evaluate PretrainZero in both the pretraining and post-training stages, showing that it effectively mitigates the general reasoning data-wall with pretraining data, and finally the pretrained reasoning models can serve as the reasoning foundation models for general downstream RLVR tasks. 
\end{itemize}

\begin{figure}[t]
\begin{center}
\centerline{\includegraphics[width=1.0\textwidth]{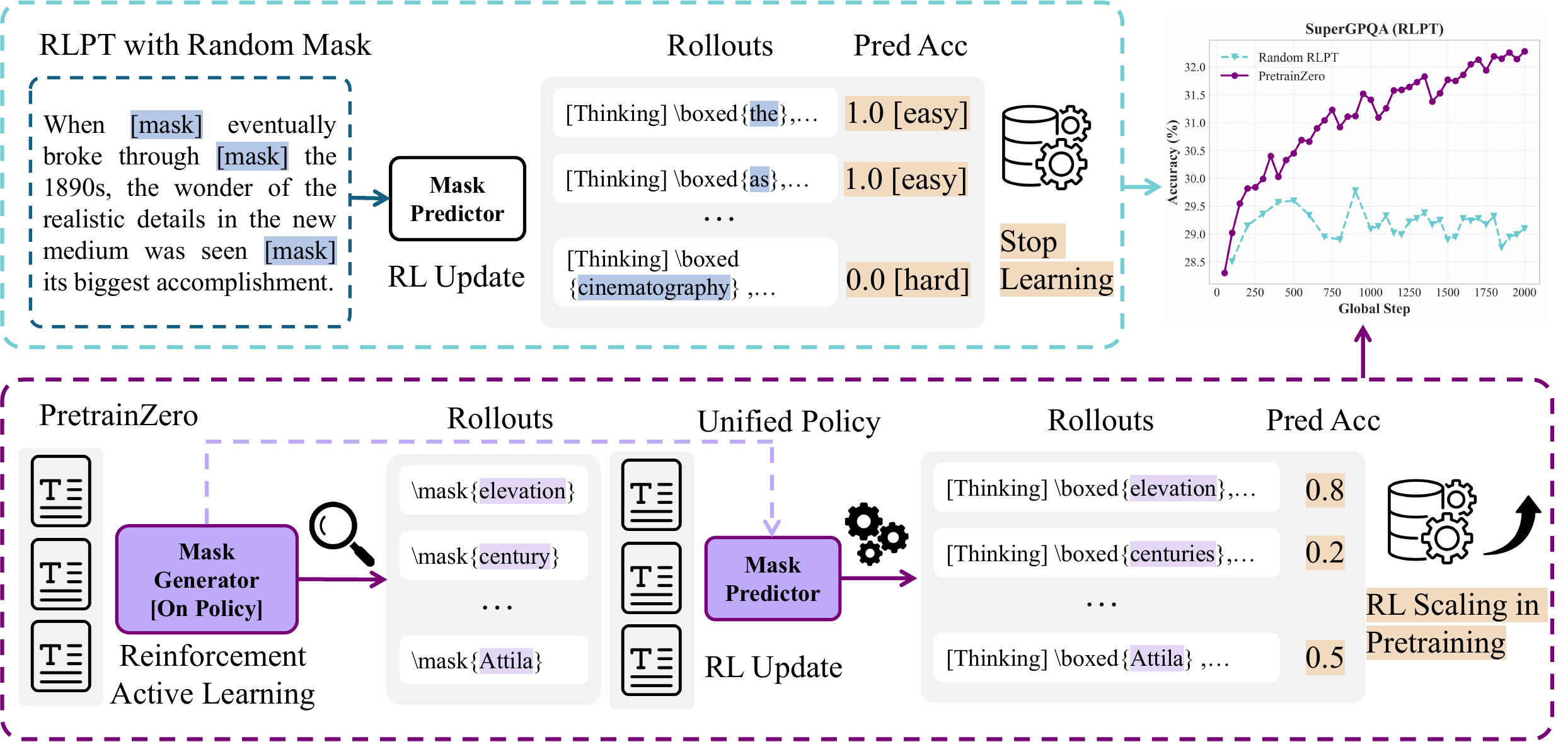}}
\vskip +0.1in 
\caption{
An overview of Reinforcement Active Pretraining. Compared with vanilla RLPT, PretrainZero actively explores and learns from the informative contexts on the pretraining corpus.}
\label{f1}
\end{center}
\vskip -0.2in 
\end{figure}

\section{Preliminary}

In order to learn from the pretraining corpus, traditional self-supervised pretraining adopts language modeling objectives to capture linguistic patterns and contextual dependencies. Recently, the emerging reinforcement pretraining constructs verifiable data through token prediction to learn the reasoning process within concepts. We briefly review different learning patterns in this section.

\subsection{Self-Supervised Pretraining}

Given the context, traditional self-supervised pretraining tasks include masked token prediction (MTP) \cite{gpt2} and next token prediction (NTP) \cite{bert}.
As shown in Eq. \ref{eq:ntp}, the NTP task predict the identity tokens $x_{t}$ in each location given their preceding context $x_{<t}$ under an auto-regressive pattern:
\begin{equation}
\label{eq:ntp}
\mathcal{J}_\text{NTP}(\theta) = \sum_{t=1}^{T} \log \pi_\theta(x_{t} \mid x_{<t}),
\end{equation}
where $x$ is the token sequence with length $T$ and $\theta$ is the pretrained model parameters. As the counterpart, masked token prediction task jointly leverages both the preceding and succeeding contexts $x_{m < t, t > n}$ to predict the masked tokens $x_{m \le t \le n}$:
\begin{equation}
\label{eq:mtp}
\mathcal{J}_\text{MTP}(\theta) = \sum_{t=m}^{n} \log \pi_\theta(x_{m \le t \le n} \mid x_{m < t, t > n}).
\end{equation}

Supported by self-supervised pretraining, modern LLMs \cite{gpt4o, liu2024deepseekv3} successfully scale up pretraining on massive Internet data. In this work, we simulate both masked token prediction and next token prediction as reinforcement reasoning tasks to explore more general RL approach \cite{ma2025general}.

\subsection{Reinforcement Pretraining}

Recent Reinforcement Pre-Training (RPT) \cite{dong2025reinforcement} extends reinforcement learning into the pretraining corpus, constructing verifiable training data directly from the pretraining corpus and thereby alleviating the reliance on costly annotations and specific environments for verification.
Specifically, RPT introduces the next-token reasoning task: given a sequence $x$, one token $x_t$ is treated as ground-truth and its preceding tokens $x_{<t}$ as context for the generated output, $o_t$.
Unlike the self-supervised NTP task, where the model directly predicts the next token, RPT \cite{dong2025reinforcement} first produces a chain-of-thought reasoning process \cite{deepseekr1} before generating the final predicted token. In optimization, RPT applies GRPO algorithm with group size $G$, and uses the exact match verifiable reward $r^{i}_t$ between prediction $\hat{x}^i_t$ and ground-truth $x_{t}^i$:

\begin{equation}
r^{i}_t = \begin{cases} 1 & \text{if } \hat{x}^i_t(x_{<t}^i) = x_{t}^i  \\ 0 & \text{otherwise} \end{cases},
\end{equation}

\begin{equation}
\label{eq:rpt}
\mathcal{J}_\text{RPT}(\theta) = \mathbb{E}_{(x_{\le t})\sim\mathcal{D},\ \{o^{i}_t\}_{i=1}^{G} \sim\pi_\theta(\cdot\mid x_{<t})}\left[r^{i}_t\right].
\end{equation}

\textbf{Discussion on the weakness of vanilla reinforcement pretraining.}
Despite the simplicity of RPT and its potential to extend the RLVR-style method into pretraining, several significant concerns also emerge, making vanilla RPT unsuitable for practical pretraining settings: 
\begin{itemize}[leftmargin=*]
\setlength\itemsep{0.01em}
\item  \textbf{Robustness on real-world corpus}: although RPT demonstrates improvements on synthetic dataset OmniMath, real-world pretraining data with more noise \cite{bert} often causes training collapsion. 
\item  \textbf{Training from base models}: vanilla RPT depends on post-training distillation; Other explorations usually rely on SFT cold-start, external reward models, significantly increasing the complexity.
\item  \textbf{Learning effectiveness}: due to the low information density in pretraining corpus, simple token selection methods fails to identify informative content, hindering effective optimization.
\item  \textbf{Training efficiency}: unlike self-supervised NTP that predicts all tokens in parallel, RPT predicts one single token in each sample, yielding limited learning information per step.

\end{itemize}

\section{Reinforcement Active Pretraining}

To solve these questions, this work first establishes a Reinforcement Learning Pre-Training (RLPT) baseline on the widely used general domain WikiPedia dataset building upon the Qwen3-4B-Base model.
Based on the empirical observations, we then propose a unified and active pretraining task to confirm the general and practical reinforcement pretraining.

\subsection{Reinforcement Pretraining Baselines}

\begin{figure}[t]
\begin{center}
\centerline{\includegraphics[width=0.99\textwidth]{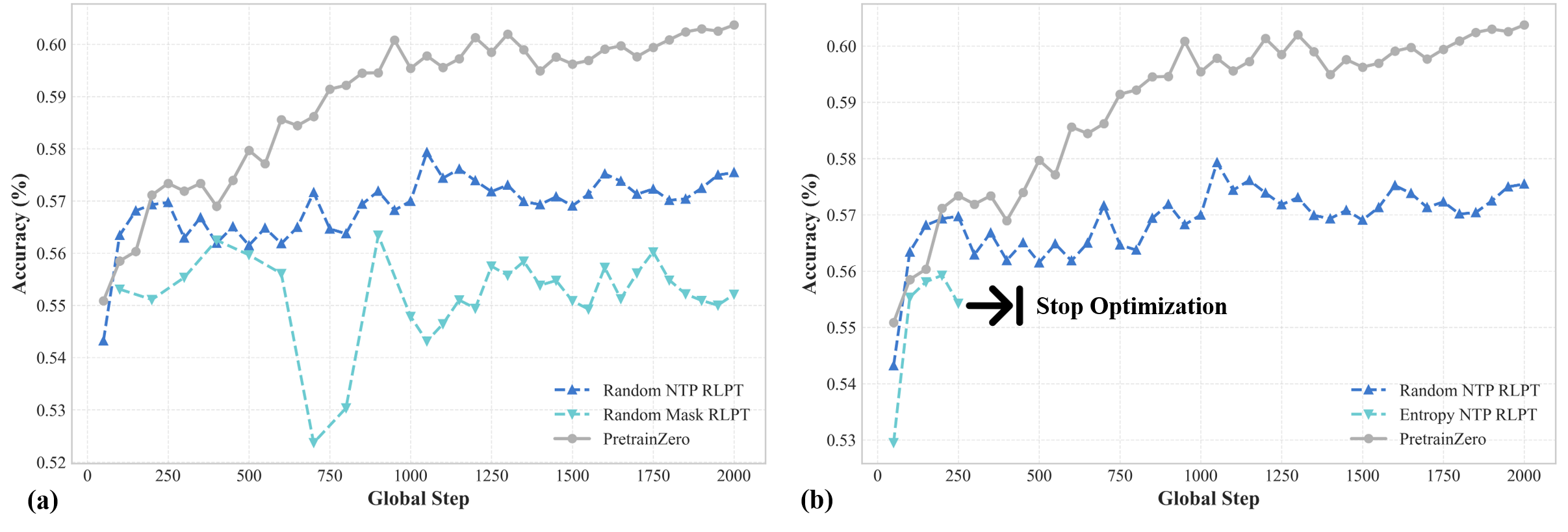}}
\caption{
MMLU-Pro performance for foundational RLPT methods. (a) Reinforcement next token prediction and reinforcement masked token prediction. (b) Reinforcement next token prediction with entropy and random token selection.}
\label{figure:f2}
\end{center}
\vskip -0.3in 
\end{figure}

We establish an RLPT baseline with three masking strategies for training corpus. The model is required to predict masked tokens $x_t$ through CoT reasoning, receiving binary rewards from exact match with ground truth:

\begin{equation}
\label{eq:rlpt}
\mathcal{J}_\text{RLPT}(\theta) = \mathbb{E}_{x\sim\mathcal{D},\ t,\ \{o^{i}_t\}_{i=1}^{G} \sim\pi_\theta(\cdot\mid x_{\backslash t})}\left[\mathbb{I}[\hat{x}_t^{i} = x_{t}^{i}]\right].
\end{equation}

Three mask prediction strategies for RLPT are investigated: 
\begin{itemize}[leftmargin=*]
\setlength\itemsep{0.01em}
\item \textbf{Random Next Token Reasoning.} The sequence is randomly truncated and the last token before truncation is masked for prediction. For each sample, the model first generates a CoT and only one selected token is predicted according to the generated CoT. 
\item \textbf{Random Masked Span Reasoning.} A word span containing several tokens in the sequence is randomly selected and masked \cite{joshi2020spanbert}, allowing the CoT to predict more than one tokens. 
\item \textbf{Entropy-based Next Token Reasoning.} The token with the top 20\% entropy in the sequence is randomly selected and masked, with all subsequent tokens truncated, which consists with RPT.
\end{itemize}

\paragraph{Empirical Observation.}
Preliminary experiments are conducted to evaluate these masking strategies on the Wikipedia corpus, with performance measured by MMLU-Pro. As shown in Figure~\ref{figure:f2} (left), Random NPT RLPT outperforms Random Mask RLPT with more stable training dynamics:

\textbf{Findings 1.} \textit{Although Mask RLPT increases the predicted tokens, the vanilla random word-span selection strategy cannot effectively capture richer semantics in pretraining.}

To further investigate the effect of token selection, Random NPT is compared with Entropy NPT, where the token with higher entropy is selected for masking. As shown in Figure~\ref{figure:f2} (right), Entropy NPT leads to training collapse and rapid reward degradation. At the position marked \textit{stop optimization}, the reward signal becomes degenerate—all samples within a group yield either 0 or 1 accuracy.
The reason is the data quality discrepancy between the synthetic and raw corpus. While entropy-based selection performs well on OmniMath (in RPT), a high-quality synthetic dataset where high-entropy tokens consistently represent challenging but learnable patterns, the same strategy fails on Wikipedia. Raw Wikipedia data contains noise and inconsistencies, causing high-entropy tokens to be either genuinely difficult or simply noisy and unpredictable, which creates unstable learning signals:

\textbf{Findings 2.} \textit{In real-world pretraining data distributions, selecting high-entropy tokens is no more effective than a random selection strategy, and learning actively from noisy data is necessary.}

\subsection{Active Pretraining Tasks}

\begin{figure}[t]
\begin{center}
\centerline{\includegraphics[width=0.8\textwidth]{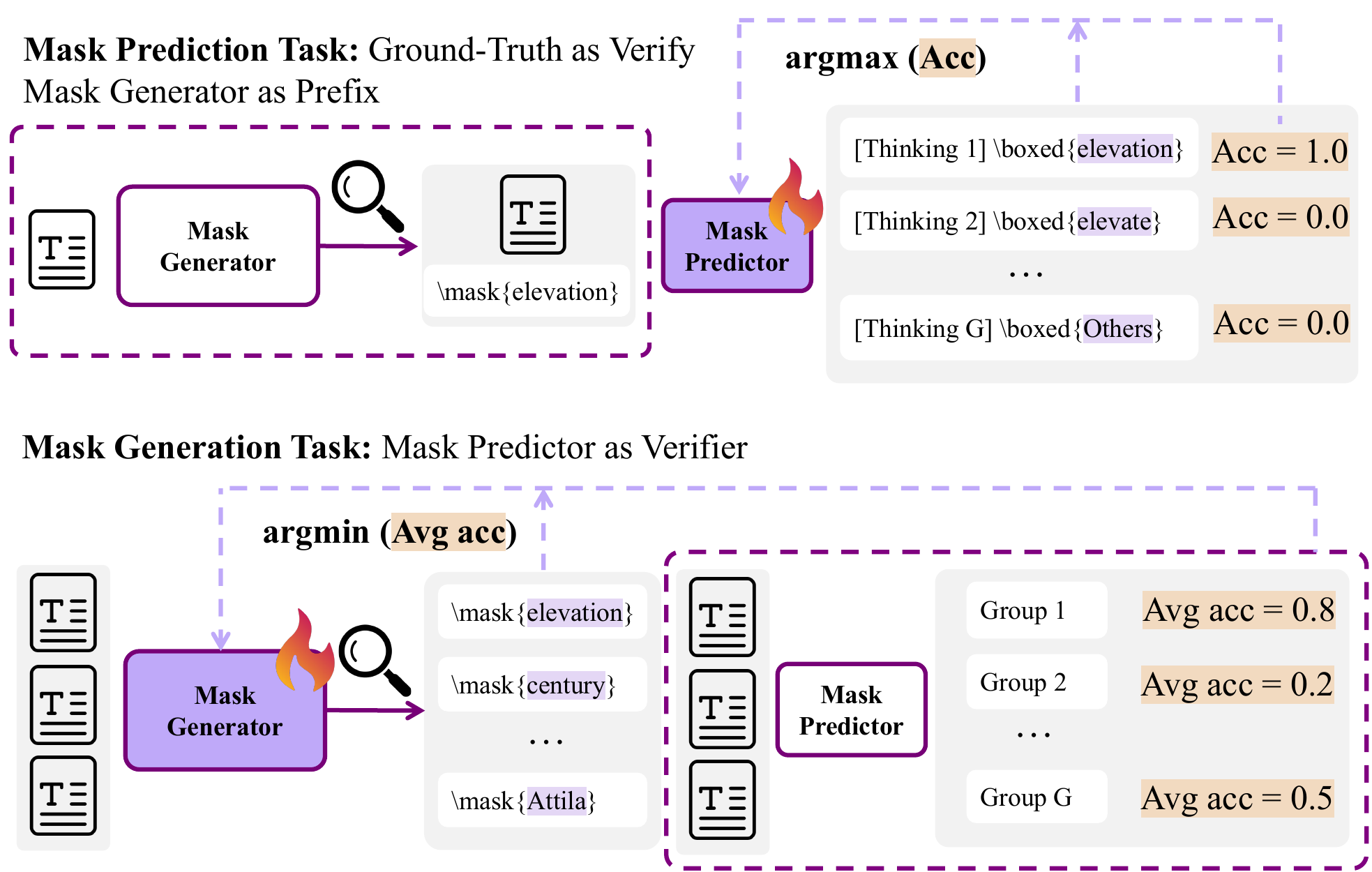}}
\caption{
Pretraining Mask Prediction and Mask Generation tasks with GRPO.}
\label{figure:f6}
\end{center}
\vskip -0.2in 
\end{figure}

\begin{figure}[t]
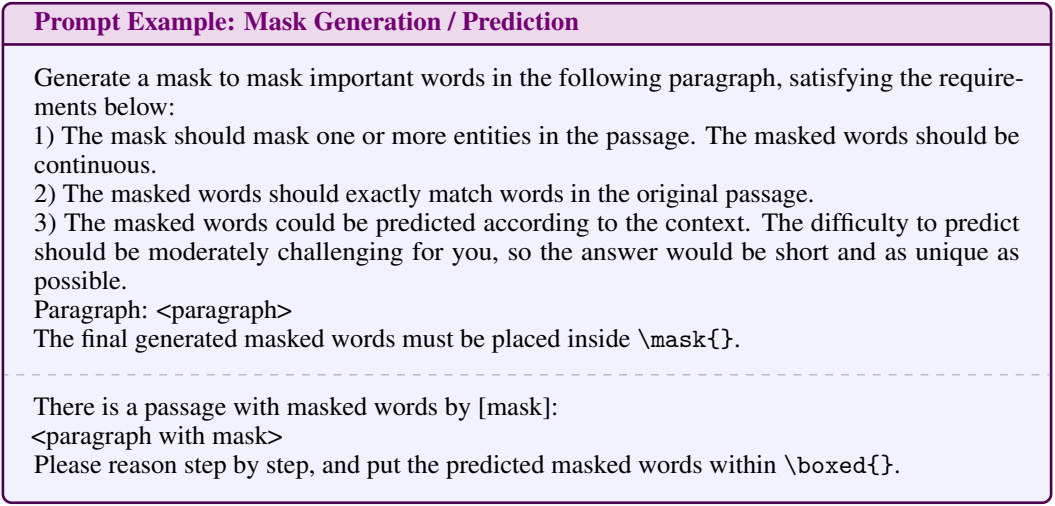

    \centering
    \begin{SplitPromptBox}[Prompt Example: Mask Generation / Prediction]
        Generate a mask to mask important words in the following paragraph, satisfying the requirements below:

1) The mask should mask one or more entities in the passage. The masked words should be continuous.

2) The masked words should exactly match words in the original passage.

3) The masked words could be predicted according to the context. The difficulty to predict should be moderately challenging for you, so the answer would be short and as unique as possible.

Paragraph: <paragraph>

The final generated masked words must be placed inside \verb|\mask{}|.
    
        \tcblower
    
        There is a passage with masked words by [mask]:

<paragraph with mask>

Please reason step by step, and put the predicted masked words within \verb|\boxed{}|.
    \end{SplitPromptBox}
    
    \caption{Prompt for Mask Generation and Prediction.}
    \label{fig:dual_prompts}
\end{figure}

The limited performance of these passive masking strategies motivates a more informative and effective learning approach. Consider how humans learn: students focus on informative and valuable content in their experience that maximizes their improvement, rather than randomly selecting practice materials. 
Inspired by the active learning behavior, we propose an active masking strategy where the model learns to identify beneficial masking positions during training. Rather than relying on fixed heuristics like random sampling or entropy thresholding, the model discovers which tokens provide the strongest learning signals. The training process consists of two tasks for the shared LLM, as shown in Fig.~\ref{figure:f6}:

\textbf{Mask Generation:} given a text sequence $s$ from pretraining data $\mathcal{D}$, the pretraining LLM $\pi_\omega$ first generates a thinking process and then generates a word span, $m \sim \pi_{\omega}(\cdot\mid s)$, to mask in this sequence. 
As shown in Fig. \ref{fig:dual_prompts}, we initially prompt the policy $\pi_{\omega}(\cdot\mid s)$ to generate a span mask with one or several words verifiable for reasoning. 
During pretraining, the policy $\pi_\omega$ continuously learns to explore and capture semantic contents from the noisy pretraining corpus by RL. 
During the early training stage, the mask prediction policy is relatively weak and requires explicit clues, while in later stages, it needs to focus on the harder and unsolved words and domains.

\textbf{Mask Prediction:} We introduce the masked span prediction as a verifiable reinforcement learning task. Different from next token reasoning, a single CoT process predicts multiple masked tokens in a continuous span, $s_{[p:q]}$. Given the generated mask $m \sim \pi_{\omega}(\cdot\mid s)$, we replace the word span $s_{[p:q]}$ with the mark $\texttt{[mask]}$ in the sequence, and then recover the masked content through CoT reasoning. 
As shown in Fig. \ref{fig:dual_prompts}, we prompt the policy $\psi_\omega(\cdot \mid m, s)$ to directly generate a CoT at the initial stage before the final mask prediction $\hat{x} \sim \psi_\omega(\cdot \mid m, s)$. 
During optimization, the CoT reasons verifiable and semantic targets from the prefixed mask generation task.

\subsection{Reinforcement Active Learning}

\textbf{Active learning objective.} We cast mask generation and mask prediction as a coupled adversarial process, implemented with a shared LLM parameterized by~$\omega$. The generator $\pi_{\omega'}(\cdot\mid s)$ proposes masking patterns, while the predictor $\psi_{\omega}(\cdot\mid m,s)$ seeks to recover the masked content. Based on the final mask prediction rewards $R(s,m,\hat{x})$, this interaction is governed by the objective: 
\begin{equation}
\label{eq:r2}
J(\omega)\;:=\;\mathbb{E}_{s\sim\mathcal{D},\;m\sim\pi_{\omega'}(\cdot\mid s)}\bigg[\,
\mathbb{E}_{\hat{x}\sim\psi_\omega(\cdot\mid m,s)}
\big[\,R(s,m,\hat{x})\,\big]\bigg],
\end{equation}
which evaluates the predictor's performance under the generator's masking strategy.
To encourage increasingly informative and challenging masks, we define the generator's
objective as $V(\omega)=\min_{\omega} J(\omega)$, while the predictor optimizes in the
opposite direction, i.e., $\arg\max_{\omega} J(\omega)$, thereby forming a coupled
min--max formulation:
\begin{equation}
\label{eq:r1}
\omega^\star = arg\max_{\omega} V(\omega) = \arg\min_{\omega' \in \Omega} 
\max_{\omega \in \Omega} 
\mathbb{E}_{s\sim \mathcal{D}}
\Big[
\mathbb{E}_{m\sim \pi_{\omega'}(\cdot\mid s),\, \hat{x} \sim \psi_\omega(\cdot \mid m, s)}
[R(s,m,\hat{x})]
\Big].
\end{equation}
This adversarial min--max structure naturally mirrors the principle of active learning, where the generator actively selects reasonable and informative masks to probe the model's weaknesses, thereby driving the predictor toward improved robustness and generalization.

\textbf{Reinforcement optimization.}
To optimize the min–max active-learning objective in Eq.~\eqref{eq:r1}, we implement both the mask prediction and mask generation as RL problems. For the prediction policy $\psi_\omega(\hat{x}\mid m,s)$, the reward is simply defined as an exact match between the predicted token span $\hat{x}^{i}$ and ground-truth $x^{i}$:
\begin{equation}
r^{i}_{\mathrm{pred}} = R(s,m,\hat{x}) = \mathbb{I}[\hat{x}^{i} = x^{i}],
\end{equation}
which directly optimizes the inner maximization $\mathbb{E}_{\hat{x} \sim \psi_\omega(\cdot \mid m, s)}[R(s,m,\hat{x})] $ in Eq.~\eqref{eq:r2}.
For the generation policy $\pi_{\omega'}(m\mid s)$, the reward is defined as the negative prediction accuracy under its own masks:
\begin{equation}
r^{j}_{\mathrm{gen}} = 1 - \mathbb{E}_{x \sim \psi_\omega(\cdot \mid m, s)}
[R(s,m,x)] = 1 - \frac{1}{G} \sum_{i=1}^{G} \mathbb{I}[\hat{x}^{i,j} = x^{i,j}],
\end{equation}
which aligns with the outer minimization $\mathbb{E}_{s\sim\mathcal{D},\;m\sim\pi_{\omega'}(\cdot\mid s)}\bigg[\,\mathbb{E}_{\hat{x}\sim\psi_\omega(\cdot\mid m,s)}\big[\,R(s,m,\hat{x})\,\big]\bigg]$ in Eq.~\eqref{eq:r2}. The mask generator is rewarded when its masks lead to lower prediction accuracy, indicating that the induced masks contain higher information content for the model. 
In addition, when the mask prediction accuracy is zero, we further define the generator’s reward to be $r^{i}_{\mathrm{gen}} = 0$, in order to avoid rewarding the noisy masks that are not predictable for $\psi_\omega(\cdot \mid m,s)$.

Given the reword definations, $r^{i}_{\mathrm{pred}}$ and $r^{j}_{\mathrm{gen}}$, we optimize Eq.~\eqref{eq:r1} using GRPO.
By substituting $r^{j}_{\mathrm{gen}}$ into the generator’s advantage, we obtain 
\( A_{\mathrm{gen}}^{j} = -\,\mathbb{E}\!\left[A_{\mathrm{pred}}^{:,j}\right] \), 
which proves that the GRPO update is fully consistent with the min--max objective in Eq.~\eqref{eq:r1}:
\begin{equation}
\label{eq:grpo0}
\hat{A}_{\mathrm{gen}}^j = \frac{r^j_{\mathrm{gen}} - \operatorname{mean}(r^1,\dots,r^G)}{\operatorname{std}(r^1,\dots,r^G)} = - \mathbb{E}[\hat{A}_{\mathrm{pred}}^{:,j}].
\end{equation}
We directly concatenate and uniformly optimize the mask generation and prediction batches in each step:
\begin{equation}
\label{eq:grpo}
\mathcal{L}_{\text{GRPO}}(\omega) = -\frac{1}{G}\sum_{i=1}^{G} \min\!\Bigl(\tfrac{\pi_\omega(x_i)}{\pi_{\omega_{\text{old}}}(x_i)}\,\hat{A}_i,\;
\operatorname{clip}\!\bigl(\tfrac{\pi_\omega(x_i)}{\pi_{\omega_{\text{old}}}(x_i)},\,1-\epsilon,\,1+\epsilon\bigr)\,\hat{A}_i
\Bigr).
\end{equation}

\section{Experimental Results}

\subsection{Implementation Details}

\textbf{Model.} To evaluate stand-alone reinforcement pretraining, we directly continue pretraining the base models using reinforcement learning without introducing any intermediate supervised finetuning (SFT) cold start. Specifically, we pretrain base models in 3 $\sim$ 30 billion parameters, including the Qwen3-4B-Base, Qwen3-8B-Base, Qwen3-30B-A3B-MoE-Base, and SmolLM3-3B-Base. 

\textbf{Dataset.} To evaluate on real-world distributed pretraining corpus, we use only the most general Wikipedia dataset. 
Notice that existing RLPT often includes explicit Question-Answer pairs or synthetic datasets such as OmniMath that contain strong reasoning CoTs; this risks allowing the RL objective to copy these reasoning traces directly, implicitly degrading to supervised learning. 

\textbf{Training.} For RLPT, we train 2000 steps using GRPO without KL regularization \cite{ppo}. Following DAPO \cite{dapo}, we filter samples whose accuracies are exactly 0.0 or 1.0, and we adopt the clip-higher strategy for stability. For Qwen-Base models, we directly perform the PretrainZero strategy; for SmolLM3-3B-Base, we first use random RLPT for 100 steps as RL cold-start, and then perform PretrainZero for the remaining 1900 steps. 
During reasoning, the max length of the prompt and response is limited to 1536 and 4096 tokens, respectively. We adopt the $5 \times 10^{-7}$ learning rate and the cosine scheme. 
In the mask-generation task, each batch contains 32 pretraining paragraphs, with 8 rollouts for each to produce masks. In the mask-prediction task, we evaluate 256 masks from the prefixed mask generation task (32$\times$8), and each mask is also paired with 8 rollouts for prediction. Consequently, the overall prompt batch size becomes 288 for RL (32 + 32$\times$8). 

\textbf{Evaluation.} We evaluate on both general-domin and math-domin reasoning benchmarks. For general domin reasoning, we evaluate on the MMLU-Pro \cite{mmlupro}, SuperGPQA \cite{supergpqa}, and BBEH \cite{kazemi2025big}; for math domin reasoning, we evaluate on 6 widely used benchmarks, including Math 500 \cite{math500}, Olympiad \cite{he2024olympiadbench}, Minerva \cite{minerva}, GSM8K \cite{gsm8k}, AMC23, and AIME24. For AMC23 and AIME24, we evaluate 32 times and report the mean@32 accuracy. We use the Qwen-Math-eval \cite{yang2024qwen2} as the math verifier.

\subsection{Pretraining Results}

\begin{figure}[t]
\begin{center}
\centerline{\includegraphics[width=1.0\textwidth]{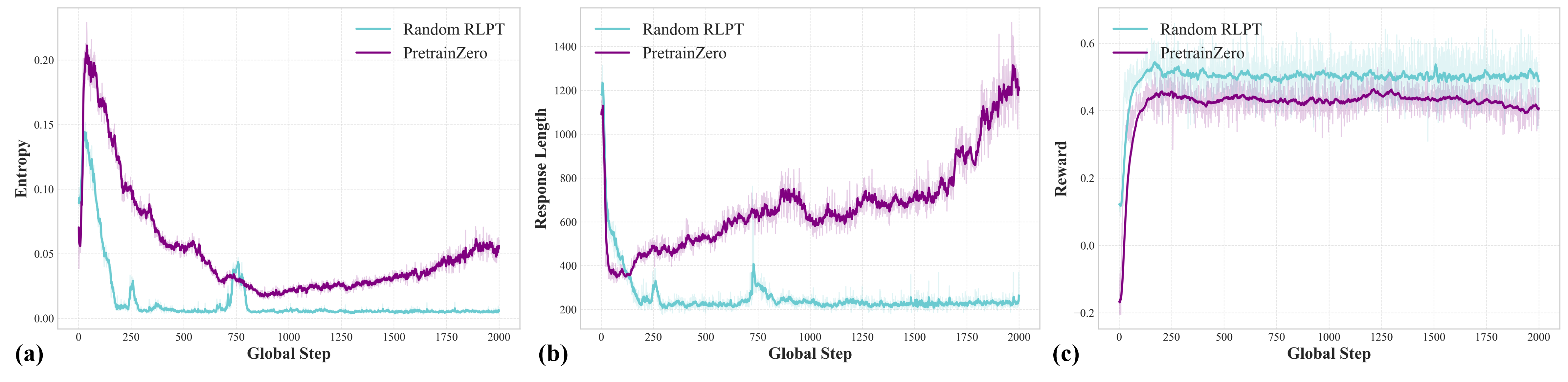}}
\caption{
Training dynamic comparisons between PretrainZero and Random RLPT on Qwen3-4B-Base: (a) entropy of model outputs; (b) response length of overall samples; (c) the overall reward.}
\label{f21}
\end{center}
\vskip -0.3in 
\end{figure}

\begin{figure}[t]
\begin{center}
\centerline{\includegraphics[width=1.0\textwidth]{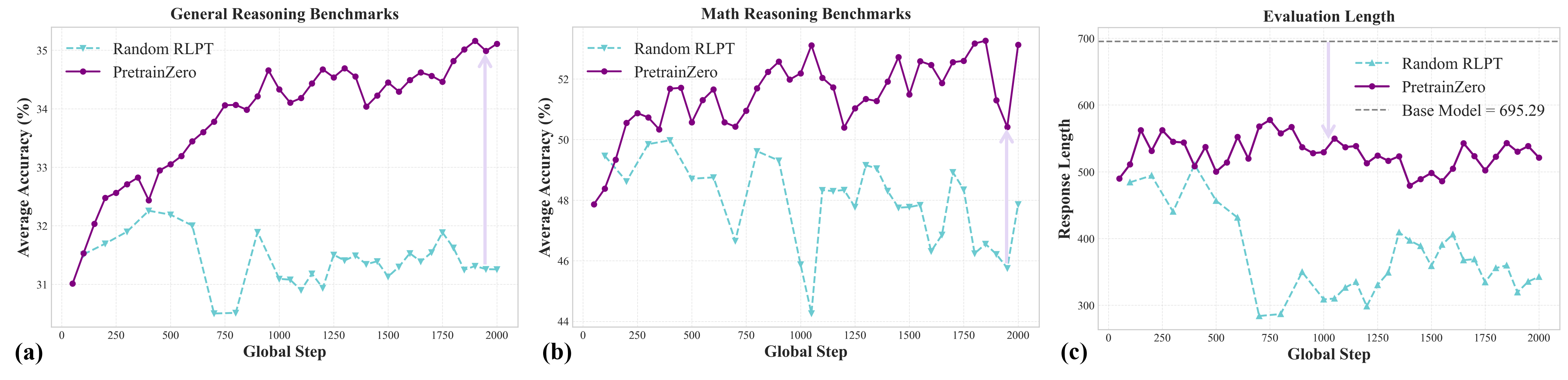}}
\vskip -0.1in 
\caption{
Evaluation comparisons between PretrainZero and Random RLPT on Qwen3-4B-Base: (a) the average accuracy on 3 general reasoning benchmarks; (b) the average accuracy on 6 math reasoning benchmarks; (c) response length on a fixed subset from MMLU-Pro.}
\label{f22}
\end{center}
\vskip -0.2in 
\end{figure}

\begin{table*}[t]
\centering
\small
\caption{Results on general-domain reasoning benchmarks. We compare the Base Model, Continue Pre-Training, Supervised Fine-Tuning, our Random RLPT baseline and PretrainZero. We highlight the best performance in \textbf{bold} and the second performance in \underline{underline}.}
\label{tab:general_domain_results}
\begin{tabular}{@{}l>{\columncolor{gray!15}}ccccc@{}}
\toprule
\textbf{Model Name} & \textbf{Overall AVG} & \textbf{MATH AVG} & \textbf{SuperGPQA} & \textbf{BBEH} & \textbf{MMLU-Pro} \\
\midrule
\multicolumn{6}{@{}l}{\textit{Qwen3-4B-Base}} \\
\quad Base Model      & 32.36 & 42.53 & 26.32 & 8.67 & 51.94 \\
\quad Continue PT  & 15.89 & 24.65 & 9.67 & 0.04 & 29.21 \\
\quad Supervised FT  &  24.27 & 15.55 & 26.38 & \underline{12.28} & 42.88 \\
\quad {Random RLPT}  & \underline{35.41} & \underline{47.87} & \underline{29.10} &{9.45} & \underline{55.21} \\
\quad \textbf{PretrainZero}  & \textbf{39.61} & \textbf{53.13} & \textbf{32.28} & \textbf{12.68} & \textbf{60.37} \\

\midrule[0.5pt]
\multicolumn{6}{@{}l}{\textit{Qwen3-8B-Base}} \\
\quad Base Model      & 37.07 & 47.48 & 31.12 & 10.49 & 59.19 \\
\quad Continue PT  & 12.32 & 27.78 & 9.94 & 0.04 & 11.51 \\
\quad Supervised FT  & 26.80 & 19.23 & 29.02 & 11.17 & 47.78 \\
\quad {Random RLPT}  & \underline{40.96} & \underline{55.08} & \underline{34.19} & \underline{12.96} & \underline{61.59} \\
\quad \textbf{PretrainZero}  & \textbf{42.78} & \textbf{57.72} & \textbf{34.46} & \textbf{14.67} & \textbf{64.28} \\

\midrule[0.5pt]

\multicolumn{6}{@{}l}{\textit{SmolLM3-3B-Base}} \\
\quad Base Model      & 16.23 & 32.31 & 12.62 & 3.32 & 16.66 \\
\quad Random RLPT     & \underline{20.25} & \underline{35.95} & \underline{14.48} & \textbf{7.85} & \underline{22.74} \\
\quad \textbf{PretrainZero}    & \textbf{23.41} & \textbf{38.03} & \textbf{19.44} & \underline{3.78} & \textbf{32.41} \\

\midrule[0.5pt]

\multicolumn{6}{@{}l}{\textit{Qwen3-30B-A3B-MoE-Base}} \\
\quad Base Model      & 38.88 & 52.49 & 33.73 & 10.51 & 58.79 \\
\quad {Random RLPT}  & \underline{40.38} & \underline{52.62} & \underline{36.33} & \underline{12.99} & \underline{59.57} \\
\quad \textbf{PretrainZero}  & \textbf{43.55} & \textbf{58.12} & \textbf{36.58} & \textbf{14.91} & \textbf{64.59} \\

\bottomrule
\end{tabular}
\end{table*}

\begin{table*}[t!]
\centering
\small
\caption{Results on math-domain reasoning benchmarks. We highlight the best performance in \textbf{bold} and the second performance in \underline{underline}.}
\label{tab:math_domain_results}
\begin{tabular}{@{}l>{\columncolor{gray!15}}ccccccc@{}}
\toprule
\textbf{Model Name} & \textbf{AVG} & \textbf{MATH-500} & \textbf{Olympiad} & \textbf{Minerva} & \textbf{GSM8K} & \textbf{AMC} & \textbf{AIME24}\\
\midrule
\multicolumn{8}{@{}l}{\textit{Qwen3-4B-Base}} \\
\quad Base Model      &  42.53 & 73.30 & 37.30 & \underline{22.10} & 86.30 & 36.17 & 0.00 \\
\quad Continue PT  &  24.65 & 38.00 & 13.60 & 11.00 & 67.00 & 15.00 & 3.30 \\
\quad Supervised FT  & 15.55 & 28.50 & 8.10 & 14.30 & 27.40 & 15.00 & 0.00 \\
\quad {Random RLPT}  & \underline{47.87} & \underline{74.80} & \underline{38.50} & \underline{22.10} & \underline{87.50} & \underline{54.30} & \underline{10.00} \\
\quad \textbf{PretrainZero}  & \textbf{53.13} & \textbf{79.10} & \textbf{42.70} & \textbf{33.80} & \textbf{92.90} & \textbf{56.95} & \textbf{13.30} \\

\midrule[0.5pt]
\multicolumn{8}{@{}l}{\textit{Qwen3-8B-Base}} \\
\quad Base Model      & 47.48 & 70.10 & 35.30 & 25.40 & 91.50 & 52.58 & 10.00 \\
\quad Continue PT  & 27.78 & 42.70 & 16.90 & 11.80 & 55.30 & 33.28 & 6.70 \\
\quad Supervised FT  & 19.23 & 30.50 & 11.70 & 15.40 & 32.80 & 25.00 & 0.00 \\
\quad {Random RLPT}  & \underline{55.08} & \underline{79.20} & \textbf{42.70} & \underline{39.00} & \underline{93.80} & \underline{62.50} & \underline{13.30} \\
\quad \textbf{PretrainZero}  & \textbf{57.72} & \textbf{81.90} & \underline{42.50} & \textbf{43.40} & \textbf{93.50} & \textbf{65.00} & \textbf{20.00} \\

\midrule[0.5pt]

\multicolumn{8}{@{}l}{\textit{SmolLM3-3B-Base}} \\
\quad Base Model      & 32.31 & 53.80 & 20.40 & 14.00 & 81.20 & 22.81 & \underline{1.65} \\
\quad Random RLPT     &\underline{35.95} & \underline{59.00} & \underline{21.50} & \underline{20.20} & \underline{82.50} &  \textbf{32.50} & 0.00 \\
\quad \textbf{PretrainZero}    &  \textbf{38.03} &  \textbf{62.60} &  \textbf{25.60} &  \textbf{22.10} &  \textbf{83.70} & \underline{27.50} &  \textbf{6.70} \\

\midrule[0.5pt]

\multicolumn{8}{@{}l}{\textit{Qwen3-30B-A3B-MoE-Base}} \\
\quad Base Model      & 52.49 & 74.70 & \underline{43.00} & 22.80 & \underline{91.10} & \underline{66.95} & \underline{16.36} \\
\quad {Random RLPT}  & \underline{52.62} & \underline{79.20} & {41.20} & \underline{38.60} & {82.40} & {59.77} & {14.58} \\
\quad \textbf{PretrainZero}  & \textbf{58.12} & \textbf{81.70} & \textbf{43.40} & \textbf{41.20} & \textbf{94.40} & \textbf{70.62} & \textbf{17.40} \\

\bottomrule
\end{tabular}
\end{table*}

\textbf{Baselines.} To compare Reinforcement Learning Pre-Training (RLPT) with conventional training patterns, we primarily establish the following baselines: 1) the base model as the initial baseline. 2) Continue Pre-Training: We continuously pretrain with the self-supervised next token prediction on the same Wikipedia data. 3) Supervised Fine-Tuning: we formulate the masked token prediction task as question-answer pairs as Fig. \ref{fig:dual_prompts}, and remove the CoT. 4) Random RLPT: We use the introduced random masked span prediction introduced in Sec. 3.1 as the strong RL baseline.

\textbf{Comparison with Supervised Pretraining.} We summarize the overall and detailed math performance in Table \ref{tab:general_domain_results} and \ref{tab:math_domain_results} respectively. Compared with the base model, Continued Pre-Training and Supervised Fine-Tuning lead to performance drops of 16.47 and 8.09 on Qwen3-4B, and 24.75 and 10.27 on Qwen3-8B, respectively. This occurs because, for highly optimized models, supervised learning on low-quality Wikipedia passages offers limited meaningful information and can even disrupt the model’s pretrained distribution.
For RLPT, Random RLPT and PretrainZero improve 3.05 and 7.25 on Qwen3-4B, and 3.89 and 5.71 on Qwen3-8B. These results suggest that reinforcement learning is capable of extracting meaningful supervision from relatively low-quality data.

\begin{figure}[t]
    \centering
    \begin{SplitPromptBoxThree}[Reasoning Patterns: Random RLPT / PretrainZero]
        \textbf{Paragraph:}  In addition to his 1983 Triple Crown wins, Ralph Hanover won seventeen additional [mask] events, including the very important Adios and Meadowlands Pace ...
    
        \tcbline
    
        \textbf{Random RLPT:} Here is the passage with the masked words filled in:

        In addition to his 1983 Triple Crown wins, Ralph Hanover won seventeen additional stakes ...

        \par\vspace{4pt}
        
        \verb|\boxed{stakes}| ...

        \tcbline

        \textbf{PretrainZero:} To determine the masked word in the sentence:

        \par\vspace{4pt}

        "In addition to his 1983 Triple Crown wins, Ralph Hanover won seventeen ..."
        
        \par\vspace{4pt}

        \textcolor{violet!90!black}{1. **Analyze the sentence structure**:} The sentence is: "Ralph Hanover won seventeen ..."
        
        \par\vspace{4pt}

        \textcolor{violet!90!black}{2. **Identify the missing word**:} The sentence structure suggests that the missing word should complete the phrase "seventeen additional [mask] events." \textcolor{violet!90!black}{Common phrases include} "seventeen additional stakes events" or "seventeen additional championship events."

        \par\vspace{4pt}
        
        \textcolor{violet!90!black}{3. **Consider the context**:} The sentence mentions specific races like ... These are all racing events, particularly in harness racing. The mention of world records and specific track distances \textcolor{violet!90!black}{further indicates that} the missing word is related to racing events.

        \par\vspace{4pt}

        \textcolor{violet!90!black}{4. **Determine the most appropriate word**:} The most common completion ... in this context is ... \textcolor{violet!90!black}{However, the most relevant term} in the context of racing achievements is "seventeen additional stakes events."

        \par\vspace{4pt}

        \textcolor{violet!90!black}{5. **Conclusion**:} The missing word is "stakes."

        Therefore, the predicted masked word is:

        \par\vspace{4pt}

        In addition to his 1983 Triple Crown wins, Ralph Hanover won seventeen additional \verb|\boxed{stakes}| events ...

    \end{SplitPromptBoxThree}
    
    \caption{Reasoning pattern comparison between Random RLPT and PretrainZero.}
    \label{fig:pattern}
\end{figure}

\textbf{Comparison with Reinforcement Pretraining.} As mentioned in Sec. 3.1, previous RPT training on high-entropy tokens quickly stops optimization when applied to real-world pretraining corpus.  We compare the training dynamics between Random RLPT and PretrainZero in Fig. \ref{f21} and \ref{f22}.
As training steps increase, we observe that both the reasoning length of PretrainZero and its performance on general- and math-reasoning benchmarks improve consistently. This indicates that PretrainZero’s reasoning ability is gradually strengthened, similar to RLVR in DeepSeek-R1.
Compared with Random RLPT, the active-learning strategy arouses longer CoT trajectories and noticeably stronger reasoning performance. As shown in Table \ref{tab:general_domain_results}, PretrainZero consistently outperforms Random RLPT by 4.20, 1.82, 3.17, and 3.16 points on the Qwen3-4B, Qwen3-8B, Qwen3-30B-A3B-MoE, and SmolLM3-3B base models, respectively.

\textbf{Reasoning Efficiency.}
As shown in Fig. \ref{f21} (b), although the growth of CoT length in training, we need not worry about the inference efficiency of the reasoning process. The growth mainly comes from improvements in the mask-prediction capability. To verify this, we sample 10\% of the MMLU-Pro prompts and evaluate the reasoning length. As shown in Fig. \ref{f22} (c), for the same questions, the reasoning length remains similar during RLPT. Moreover, compared with the base model, RLPT actually improves the efficiency of CoT reasoning.

\textbf{Reasoning Pattern.}
As shown in Fig. \ref{fig:pattern}, we compare the reasoning patterns between Random RLPT and PretrainZero from the Qwen3-8B-Base. Given the same masked target, Random RLPT directly outputs the answer without any explicit reasoning. In contrast, PretrainZero first explores multiple possibilities, analyzes and verifies them step by step, and finally summarizes to reach a conclusion. 
Since the mask-prediction objective does not appear in downstream RL tasks, the emergence of such reasoning behavior during pretraining provides a stronger reasoning ability for generalization.

\begin{figure}[t]
\begin{center}
\centerline{\includegraphics[width=0.98\textwidth]{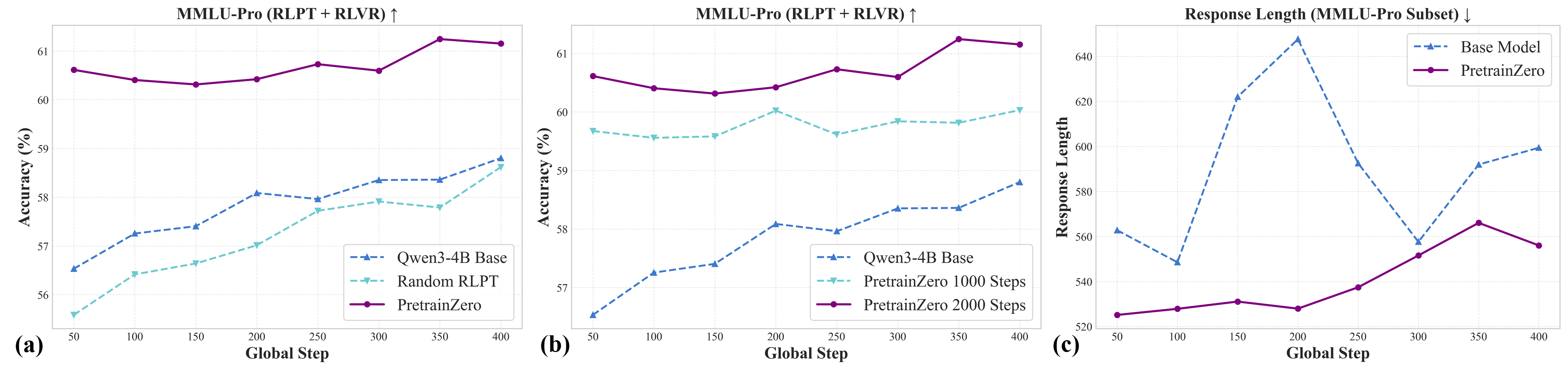}}
\caption{
RLPT performance after the same RLVR post-training. 
(a) Comparison of Qwen3-4B-base, Random RLPT, and PretrainZero.
(b) Comparison of Qwen3-4B-base, PretrainZero with 1000 and 2000 steps RLPT.
(c) Response length comparison in the same MMLU-Pro subset.}
\label{f7}
\end{center}
\vskip -0.3in 
\end{figure}

\subsection{Post-Training Results}

To investigate whether PretrainZero can improve the general reasoning capabilities of the foundation model for efficient RL finetuning, we apply RLVR as a post-training stage on PretrainZero. 
For the general RLVR task, we follow the General Reasoner recipe \cite{ma2025general}. Specifically, we apply the Web-Instruct dataset \cite{ma2025general} in a Question–Answer format, and the same pretrained reward model as the verifier.
For efficient RL finetuning, we train 400 steps on Qwen3-4B series models with a single node with $8\times$ H800 GPUs, which supports at most the 128 batchsize, 1/8 compared with General Reasoner.

We evaluate the training process in Fig.~\ref{f7}, and report the final general-domain and math-domain performance in Table~\ref{tab:general_post} and Table~\ref{tab:math_post}, respectively.
As shown in Fig.~\ref{f7} (b), the performance consistently improves as the training starts progressing from the base model to PretrainZero at 1000 RLPT steps and further to PretrainZero at 2000 RLPT steps on MMLU-Pro. 
As shown in Fig.~\ref{f7} (c), PretrainZero has more stable and efficient CoT in downstream RLVR.
Compared with the base model, PretrainZero significantly improves the math average and overall accuracy by 2.18 and 2.56 points end-to-end.

\begin{table*}[t]
\centering
\small
\caption{Results on general-domain reasoning benchmarks after the RLVR post-training. We perform the general RLVR post-training \cite{ma2025general} from the Qwen3-4B-Base model, Random RLPT, and PretrainZero with 1000 / 2000 step RLPT. \textbf{RLPT / RLVR} indicates RL steps in RLPT and RLVR stages respectively. We highlight the best performance in \textbf{bold}.}
\label{tab:general_post}

\resizebox{\textwidth}{!}{
\begin{tabular}{@{}lc>{\columncolor{gray!15}}ccccc@{}}
\toprule
\textbf{Model Name} &\textbf{RLPT / RLVR} & \textbf{Overall AVG} & \textbf{MATH AVG} & \textbf{SuperGPQA} & \textbf{BBEH} & \textbf{MMLU-Pro} \\
\midrule
\quad Base Model  &-- / 400    & 37.90 & 50.96 & 30.26 & 11.59 & 58.80 \\
\quad Random RLPT &2000 / 400 & 38.43 & 51.49 & 30.77 & 12.83 & 58.62 \\
\quad PretrainZero &1000 / 400 & 39.15 & 51.84 & 32.32 & 12.39 & 60.03 \\
\quad \textbf{PretrainZero} & {2000} / 400 &  \textbf{40.46} & \textbf{53.77} & \textbf{33.30} & \textbf{13.61} & \textbf{61.15} \\
\bottomrule
\end{tabular}
}
\end{table*}

\begin{table}[t]
\centering
\small
\caption{Results on math-domain reasoning benchmarks after the RLVR post-training. \textbf{RLPT / RLVR} indicates RL steps in RLPT and RLVR stages respectively. }
\label{tab:math_post}

\resizebox{\columnwidth}{!}{
\begin{tabular}{@{}lc>{\columncolor{gray!15}}ccccccc@{}}
\toprule
\textbf{Model Name} & \textbf{RLPT / RLVR} &\textbf{AVG} & \textbf{MATH-500} & \textbf{Olympiad} & \textbf{Minerva} & \textbf{GSM8K} & \textbf{AMC} & \textbf{AIME24}\\
\midrule
\quad Base Model    &-- / 400    & 50.96 & 75.70 & 41.80 & 31.60 & 91.30 & 52.03 & \textbf{13.30} \\
\quad Random RLPT   &2000 / 400  & 47.87 & 74.80 & 38.50 & 22.10 & 87.50 & 54.30 & 10.00 \\
\quad PretrainZero  &1000 / 400  & 51.84 & 77.00 & 43.00 & 32.40 & 92.50 & 55.00 & 11.13 \\
\quad \textbf{PretrainZero}  & {2000} / 400  & \textbf{53.77} & \textbf{78.80} & \textbf{43.00} & \textbf{39.70} & \textbf{93.00} & \textbf{54.84} & \textbf{13.30} \\

\bottomrule
\end{tabular}
}
\end{table}

\subsection{Ablation Studies}

\begin{figure}[h]
\begin{center}
\centerline{\includegraphics[width=0.99\textwidth]{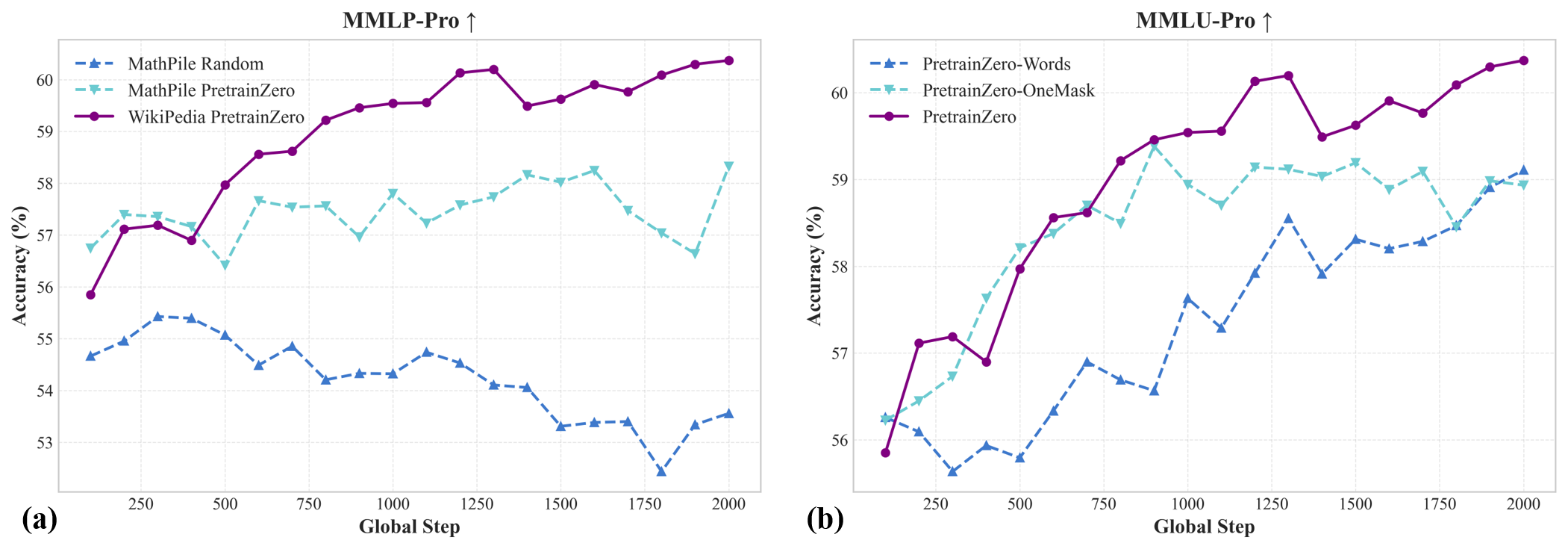}}
\caption{
Comparisons for data domain and mask regularization. (a) MMLU-Pro performance on MathPile and Wikipedia. (b) MMLU-Pro performance with different mask regularization strategies.}
\label{f8}
\end{center}
\vskip -0.3in 
\end{figure}

\textbf{Specific Domain.} 
To explore the impact of data domain on RLPT, we compare RLPT performance on the Wikipedia corpus in the general-domain versus the MathPile \cite{wang2024mathpile} dataset in the math-domain.
As shown in Fig.~\ref{f8} (a), directly using general-domain Wikipedia data yields better performance. Since curating high-quality mathematical data requires substantial expert effort, we recommend using general-domain data for a much lower cost of data acquisition.

\textbf{Training Robustness.}
To confirm RLPT over 2000 steps, we evaluate different mask regularization strategies:
1) PretrainZero: For the generated masks, we retain only those whose underlying spans appear fewer than eight times within the paragraph.
2) PretrainZero-OneMask: Based on PretrainZero, if a generated mask appears multiple times in the paragraph, we randomly replace only one occurrence with \texttt{[mask]} and make prediction.
3) PretrainZero-Words: Since PretrainZero may produce masks that cover incomplete words—reducing interpretability—we filter masks that keep only complete word spans.
As shown in Fig.~\ref{f8} (b), three recipes can be trained stably, and PretrainZero consistently achieves better performance.

\section{Related Works and Discussion}

\textbf{Self-Supervised Pretraining for LLMs.} 
Scalable self-supervised pretraining \cite{scaling:law} formulates the foundation of advanced large language models. Under the simple and fixed learning pattern, the next token prediction, autoregressive LLMs \cite{transformer, team2025kimi} can be trained on massive corpus at the Internet-scale, establishing strong general-purpose capabilities. Beyond this pattern, token-masked prediction objectives \cite{bert, spanbert} continue to play an important role in the pretraining of language models, such as in BERT-style embedding models \cite{chen2024bge}, diffusion language models \cite{nie2025large}, and code-focused pretraining \cite{hui2024qwen2}. The reliability, scalability, and broad applicability of self-supervised learning offer key insights for reinforcement pretraining and highlight its potential as a fundamental training strategy.

\textbf{Reinforcement Learning for LLMs.} 
Recent large reasoning models are largely driven by post-training reinforcement learning, enabling human-expert performance in specialized domains such as web agents \cite{team2025tongyi}, tool use \cite{patilberkeley}, software development \cite{jimenez2023swe}, and mathematics \cite{deepseekr1}. Despite this progress, existing RLHF \cite{instructgpt, bai2022training} and RLVR \cite{tulu3} approaches rely heavily on human annotation and domain-specific verification environments, leading to a severe data bottleneck in general domains \cite{ma2025general, zhou2025reinforcing}. For RLHF, reward models must be continuously updated with human-labeled data to avoid reward hacking. For RLVR, training data must come from domains with verifiable ground-truth answers, and the construction of verifiable environments fundamentally limits its scalability for general reasoning tasks \cite{ma2025general}. 

\textbf{Reinforcement Pretraining.} 
To overcome the substantial verification data-wall, Reinforcement Learning Pre-Training (RLPT) has recently emerged as a promising direction, which constructs general-purpose RLVR directly on pretraining corpus using self-supervised objectives. 
Early works including Quiet-STaR \cite{zelikman2024quiet} and Fast Quiet-STaR \cite{huang2025fast} focus on token-level reasoning.
Reinforcement Pre-Training (RPT) \cite{dong2025reinforcement} is the first to apply the next-token–prediction as the RLVR objective, demonstrating the feasibility of general-purpose RL. However, RPT remains limitations, such as relying on synthetic OmniMath data with CoT annotations rather than real pretraining distributions, and training from a post-trained model instead of a base model, which prevents RPT from being practical and prolonged \cite{liu2025prorl} RLPT.

Recently, PRT \cite{hatamizadeh2025rlp} and RLPT$^{1}$ \cite{li2025reinforcement} are proposed around the similar period as this work. {PRT} incorpustes reinforcement learning as an auxiliary objective to the self-supervised pretraining and does not exclude some QA-style training data. RLPT$^{1}$ employs an additional reward model as a verifier for the sentence-level prediction objective and further introduces a high-quality SFT cold-start. 
Despite these advantages, the foundational questions in RLPT remain unexplored: under fully self-supervised conditions—removing reward models, SFT cold start, and supervised cross-entropy losses—can stand-alone RLPT be effectively trained on noisy, real-world pretraining corpus? And how to improve learning efficiency in low-information-density pretraining data? Addressing these fundamental questions becomes the primary focus of this work.

\section{Conclusion}

This work introduces the stand-alone reinforcement pretraining method in a real-world pretraining corpus, named PretrainZero. Coupled with PretrainZero, a new reinforcement active pretraining framework is proposed to explore informative, verifiable, and not-yet-mastered content in noisy pretraining data. Thanks to active learning ability, PretrainZero significantly surpasses previous fixed learning patterns, such as continued pretraining, supervised fine-tuning, and random or entropy-based reinforcement pretraining. We reveal that even Wikipedia, which has already been trained during base model pretraining, can successfully improve end-task performance with reinforcement and active learning methods. We believe that there would be great potential to explore more efficient learning patterns to discover latent information from the pretraining corpus in the future.

\bibliography{example_paper}
\bibliographystyle{alpha}

\end{document}